\documentclass[twoside]{article}

\usepackage[preprint]{aistats2027}

\usepackage[round]{natbib}

\usepackage[utf8]{inputenc}
\usepackage[T1]{fontenc}
\usepackage{hyperref}
\usepackage{url}
\usepackage{booktabs}
\usepackage{amsfonts}
\usepackage{nicefrac}
\usepackage{microtype}
\usepackage{xcolor}
\usepackage{float}
\usepackage{mathtools}
\usepackage{amssymb}
\usepackage{graphicx}
\usepackage{subfigure}
\usepackage{amsthm}
\usepackage{multirow}
\usepackage{longtable}
\newtheorem{theorem}{Theorem}
\newtheorem{lemma}{Lemma}
\newtheorem{proposition}{Proposition}
\newtheorem{assumption}{Assumption}[section]
\allowdisplaybreaks

\begin{document}

\runningauthor{Renjie Cao, Yifei Cheng, Li Shen}

\twocolumn[

\aistatstitle{Formalizing Forgetting: Theory of Forgetting for Online Continual Learning}

\aistatsauthor{ Renjie Cao \And Yifei Cheng \And Li Shen$^{\dagger}$ }

\aistatsaddress{ Boston College \\ {\small\texttt{caoren@bc.edu}}
  \And Sun Yat-sen University \\ {\small\texttt{chengyf28@mail.sysu.edu.cn}}
  \And Sun Yat-sen University \\ {\small\texttt{mathshenli@gmail.com}} } ]

{\renewcommand{\thefootnote}{}\footnotetext{$^{\dagger}$Corresponding author.}}

\begin{abstract}
  Online continual learning (OCL) requires a learner to acquire a sequence of tasks in a single pass over the data, and catastrophic forgetting of earlier tasks remains its central challenge. Despite the many methods proposed to mitigate forgetting, the theoretical understanding of forgetting in OCL is still limited: existing analyses mostly focus on linear, overparameterized, or neural tangent kernel regimes, and it remains unclear which factors govern forgetting and how they affect it. To fill this gap, we analyze single-pass Shuffle SGD under strongly convex and general convex objectives and derive explicit upper bounds on the expected final forgetting. Further analysis of these bounds shows how the per-task data amount, task similarity, and task ordering enter them, and that their effects can differ across objective classes. Moreover, experiments with neural networks on Split-CIFAR-100 and PMNIST show that the trends suggested by our theory also appear in practice, indicating that these insights extend beyond the convex setting.
\end{abstract}

\section{Introduction}\label{sec:intro}

In online continual learning (OCL), a learner is trained on a non-stationary stream of tasks and observes each sample only once~\citep{aljundi2019gradient}. The central challenge is to balance plasticity, the ability to learn new tasks, with stability, the ability to retain previously learned ones. When the model adapts too aggressively to incoming data, its performance on earlier tasks can degrade severely, a phenomenon known as catastrophic forgetting~\citep{MCCLOSKEY1989109}. Forgetting is particularly pronounced in the online setting, because the single-pass constraint prevents the learner from revisiting past samples, which makes forgetting both more likely and harder to mitigate.

Many methods have been proposed to address this stability--plasticity dilemma. Rehearsal-based methods maintain a small buffer of past samples~\citep{md2023grasp, wu2024mitigating} or replay generated features of past tasks~\citep{li2025feature} to consolidate old knowledge. Other approaches protect useful parameters through utility-based gradient perturbation~\citep{mohamed2024addressing} or use self-distillation~\citep{kotaro2024reducing} to preserve past representations. More recently, parameter-efficient fine-tuning with online low-rank adaptation (LoRA)~\citep{xiwen2024onlinelora} and contrastive replay with dual margins~\citep{lyu2025mitigating} have been used to reduce interference between tasks. In parallel, theoretical work has begun to characterize forgetting dynamics; for instance, \citet{haozhe2024order} develop a statistical-mechanics theory that identifies phase transitions of continual learning in deep networks. Nevertheless, analytical frameworks that express the expected forgetting explicitly in terms of interpretable factors remain scarce.

Understanding which factors govern forgetting is therefore essential, yet such analyses are still rare. The work most closely related to ours is \citet{lin2023theory}, which derives exact expressions for the expected forgetting of overparameterized linear models and highlights the roles of overparameterization, task similarity, and task ordering. We pursue the same goal of making forgetting analytically interpretable, but study single-pass Shuffle SGD under general convex and strongly convex objectives. This distinction matters because the mechanisms that govern forgetting change with the contraction or growth properties of the objective: in our analysis, data amount, task similarity, and task ordering remain the central factors, but their effects are mediated by the objective class through the coefficients of the bounds.

Our theoretical results are as follows. (1) \emph{Data amount}: for strongly convex objectives, increasing the per-task data amount $n$ decreases the propagation factor of the bound, so the task-gap part of the bound decays exponentially in $n$, while its variance part grows linearly in $n$ through the shuffling variance; for general convex objectives, $n$ enters only through an accumulated variance term, which grows with $n$ within the stepsize range of the analysis. (2) \emph{Task similarity}: the gap between task optima enters the bounds through nonnegative, order-dependent coefficients, so a larger gap never decreases the bound, while its weight depends on where the two tasks appear in the sequence. (3) \emph{Task ordering}: the coefficients favor spacing similar tasks apart when the propagation factor of the strongly convex bound is below one, and placing them close to each other otherwise, including the general convex case. Experiments with MLPs on Split-CIFAR-100 and PMNIST are consistent with the qualitative role of these factors. In addition, we observe empirically that less data per task is associated with higher accuracy and backward transfer.

In summary, our contributions are threefold. (1) For OCL with convex objectives, we derive explicit upper bounds on the expected forgetting that make the governing terms directly visible. (2) We show how data amount, task similarity, and task ordering interact to shape these bounds, and that the resulting guidance differs across objective classes. (3) We provide empirical evidence on standard continual-learning benchmarks that supports the qualitative implications of the theory.

\section{Related work}

\paragraph{Empirical methods for OCL.}
Most OCL methods rely on rehearsal, regularization, or the adaptation of pre-trained models. Rehearsal-based methods select which samples to store or replay~\citep{aljundi2019gradient, md2023grasp}, stabilize representations and optimization under replay~\citep{lucas2021new, yoo2024layerwise}, model the relations among previous tasks~\citep{wu2024mitigating}, or select the knowledge relevant to the current task~\citep{yanan2022selecting}. Other approaches protect useful parameters~\citep{mohamed2024addressing}, distill past knowledge~\citep{kotaro2024reducing}, average the model with earlier checkpoints~\citep{anat2025soup}, or adapt pre-trained models through low-rank adaptation and prompts~\citep{xiwen2024onlinelora, wang2025rehearsalfreetaskfreeonlinecontinual}. Empirical studies further examine how model size~\citep{eunhae2024the}, limited computation~\citep{zhipeng2022improving}, synthetic data contamination~\citep{wang2024dealing}, and model throughput~\citep{xinrui2024forgetting} affect OCL, and extend it to embodied agents~\citep{kim2024online}. In contrast to these algorithmic and empirical studies, we ask how properties of the task stream itself govern forgetting.

\paragraph{Theoretical analysis of forgetting.}
Most theoretical analyses consider linear models, where training on a task admits a closed-form description. When an overparameterized linear model is fitted to each task, \citet{evron2022catastrophic} connect continual learning to alternating projections and bound the forgetting under cyclic and random task orderings; \citet{swartworth2023nearly} tighten the bound for cyclic orderings, and \citet{evron2025sgd} show that fitting a task is equivalent to an SGD step on a modified objective and derive dimension-independent rates for random orderings. \citet{lin2023theory} obtain exact expressions for the expected forgetting and generalization error of overparameterized linear models with Gaussian features, \citet{goldfarb2023analysis} study two tasks related by a random orthogonal transformation, and \citet{meng2024understanding} analyze SGD in both the underparameterized and overparameterized regimes. \citet{itay2023continual} extend this line of work to separable linear classification. Beyond linear models, \citet{doan2021theoretical} analyze forgetting in the NTK regime through the overlap between task kernels, \citet{rathin2025pathcoordinated} derive NTK-based plasticity bounds, and \citet{haozhe2024order} develop a statistical-mechanics theory for deep, wide networks. Other works analyze specific mechanisms, such as mixture-of-experts models~\citep{hongbo2024theory}, the Ideal Continual Learner~\citep{liangzu2023the}, Bayesian filtering for non-stationary classification~\citep{titsias2024kalman}, and replay~\citep{junze2025unlocking, mohammadamin2024theoretical}, which can even increase forgetting in overparameterized linear models~\citep{yasaman2025replay}.

\paragraph{Task similarity and task ordering.}
The dependence of forgetting on the relations between tasks is subtle. \citet{ramasesh2021anatomy} find that forgetting is maximal for task sequences of intermediate similarity, \citet{lee2021continual} obtain the same effect analytically in a teacher--student setup when tasks depend on similar features, and \citet{doan2021theoretical} show that forgetting increases as two tasks become more aligned in the NTK regime. In overparameterized linear models, \citet{lin2023theory} show that forgetting need not increase with the gap between task optima, because the corresponding coefficient can be negative, and that alternating dissimilar tasks minimizes forgetting in special cases; \citet{evron2022catastrophic} and \citet{meng2024understanding} further show that the task order affects forgetting.

\paragraph{Comparison with prior analyses.}
Our analysis differs from the linear analyses above in three respects. First, the objectives are general convex or strongly convex rather than least squares, which covers losses such as regularized logistic regression. Second, each task is learned by a single pass of Shuffle SGD rather than being fitted to convergence or solved in closed form, so the per-task data amount $n$ enters through both the contraction of the optimizer and its stochastic noise. Third, because the available Shuffle-SGD guarantees are inequalities, we obtain upper bounds rather than the exact expressions of \citet{lin2023theory}. These differences change the conclusions. In our bounds, the gap between two tasks always enters with a nonnegative weight, so the non-monotone effects of similarity reported by \citet{ramasesh2021anatomy}, \citet{lee2021continual}, and \citet{lin2023theory} are not captured by an upper bound. Moreover, the order favored by the task-gap terms of our bounds depends on whether their propagation factor is below one: spacing similar tasks apart is favored in this case, in agreement with the alternating orders of \citet{lin2023theory}, and clustering them is favored otherwise.

\section{Preliminaries}

\label{sec:preliminary}

\paragraph{Problem Setting.}
We consider the standard OCL setup in which a sequence of $T$ tasks arrives over time, indexed by $\mathcal{T}=\{1,\ldots,T\}$.
For each task $t\in\mathcal{T}$, the learner observes a dataset
$S_t=\{z_{t,1},\ldots,z_{t,n}\}$, drawn i.i.d.\ from a distribution $\mathcal{D}_t$, and aims to minimize the finite-sum objective
\begin{equation}
\begin{split}
    f_t(w)\;&:=\;\frac{1}{n}\sum_{i=1}^n f_{t,i}(w), \\
    f_{t,i}(w)&:=\ell(w;z_{t,i}),\;\; w\in\mathbb{R}^d.
\end{split}
\end{equation}
Let $w_t^*\in\arg\min_w f_t(w)$ denote an optimal solution of task $t$.
We write $w_t$ for the parameter obtained after training on task $t$, which initializes training on task $t+1$, and set the starting point to $w_0=0$.
Unlike offline continual learning, each task is processed once as it arrives, and the learner cannot revisit the data of previous tasks unless a replay mechanism is introduced; we study the setting without replay.

\begin{assumption}
\label{assump:smooth-convex}
For all $t\in\mathcal{T}$ and $i\in\{1,\ldots,n\}$, the component $f_{t,i}$ is $L$-smooth.
We consider two regimes: (i) \emph{general convex}, where every $f_{t,i}$ is convex, and (ii) \emph{strongly convex}, where every $f_{t,i}$ is $\mu$-strongly convex for some $\mu>0$.
These component-wise conditions are required by the Shuffle-SGD analysis of \citet{mishchenko2021randomreshufflingsimpleanalysis} used below, and they imply the same properties for $f_t$.
\end{assumption}

\begin{assumption}
\label{assump:variance}
Following \citet{mishchenko2021randomreshufflingsimpleanalysis}, we use the variance at the optimum
\begin{equation}
    \sigma_*^2 \;:=\;\frac{1}{n}\sum_{i=1}^n \bigl\|\nabla f_{t,i}(w_t^*)\bigr\|^2,
\end{equation}
which we assume to be finite and uniformly bounded across tasks. In the strongly convex case, we also use the shuffling variance $\sigma_S^2$ of \citet{mishchenko2021randomreshufflingsimpleanalysis}, defined in Appendix~\ref{app:notation}.
\end{assumption}

\begin{assumption}
\label{assump:equal}
For simplicity, all tasks have the same number of training samples and the same noise level,
i.e., $n_t = n$ for all $t\in\mathcal{T}$, and $\sigma_*$ and $\sigma_S$ are the same for all tasks.
\end{assumption}

\paragraph{Learning Procedure.}
For each task $t$, we run one pass (one epoch) of Shuffle SGD~\citep{mishchenko2021randomreshufflingsimpleanalysis} over $S_t$: we sample a random permutation $\pi_t$ of $\{1,\ldots,n\}$, set $w_t^{(0)}:=w_{t-1}$, and for $k=0,\ldots,n-1$ update
\begin{equation}
    w_t^{(k+1)} \;=\; w_t^{(k)} \;-\;\gamma\, \nabla f_{t,\pi_t(k)}\!\left(w_t^{(k)}\right),
\end{equation}
and output $w_t:=w_t^{(n)}$.
Once task $t$ is finished, its data $S_t$ is not revisited.
Our analysis therefore concerns a single-pass Shuffle-SGD learner, not an offline learner that repeatedly trains on the accumulated data.

\paragraph{Performance Evaluation.}
To quantify forgetting theoretically, we measure the deviation of the current model from the optimal solutions of previous tasks.
We define the model error on task $i$ as
\begin{equation}
    K_i(w)\;:=\;\|w-w_i^*\|^2.
\end{equation}
Following \citet{lin2023theory}, the forgetting after learning task $t$ is defined as
\begin{equation}
\label{eq:forgetting_def}
    F_t \;:=\;\frac{1}{t-1}\sum_{i=1}^{t-1}\Bigl(K_i(w_t)-K_i(w_i)\Bigr),
\end{equation}
and we focus on the expected final forgetting $\mathbb{E}[F_T]$.
This parameter-distance quantity is a theoretical surrogate for forgetting that makes the analysis tractable under convex objectives. In the experiments, we report standard continual-learning metrics, namely final accuracy and backward transfer (BWT).

\section{Theoretical Results}
\label{r4}

This section analyzes the expected final forgetting $\mathbb{E}[F_T]$. By Eq.~\eqref{eq:forgetting_def}, it involves the distance $\|w_t-w_i^*\|^2$ between the continually updated model and the optima of previous tasks. To connect the optimization trajectory with cross-task forgetting, we first state the convergence of Shuffle SGD within a single task.

\begin{proposition}[{\citealp[Theorem~1]{mishchenko2021randomreshufflingsimpleanalysis}}]
\label{prop:intra_strong}
Under Assumptions~\ref{assump:smooth-convex} and~\ref{assump:variance} in the strongly convex regime with stepsize $\gamma\le 1/L$, the parameter $w_t$ obtained after one pass of Shuffle SGD on task $t$, starting from $w_{t-1}$, satisfies
\begin{equation}
\label{eq:prop_strong}
\mathbb{E}[\|w_t - w_t^*\|^2] \leq (1 - \gamma\mu)^n \mathbb{E}[\|w_{t-1} - w_t^*\|^2] + \frac{2\gamma\sigma_S^2}{\mu}.
\end{equation}
\end{proposition}

\begin{proposition}[{\citealp[Lemma~3]{mishchenko2021randomreshufflingsimpleanalysis}}]
\label{prop:intra_general}
Under Assumptions~\ref{assump:smooth-convex} and~\ref{assump:variance} in the general convex regime with stepsize $\gamma\le 1/(\sqrt{2}Ln)$, one pass of Shuffle SGD on task $t$ satisfies
\begin{equation}
\label{eq:prop_general}
\begin{aligned}
\mathbb{E}[\|w_t - w_t^*\|^2] \leq\ & \mathbb{E}[\|w_{t-1} - w_t^*\|^2] + \frac{\gamma^3 L n^2 \sigma_*^2}{2} \\
& \quad - 2\gamma n \mathbb{E}[f_{t}(w_t) - f_{t}(w_t^*)].
\end{aligned}
\end{equation}
\end{proposition}

These propositions only control the distance to the current optimum $w_t^*$, whereas the forgetting of a past task $i<t$ depends on the distance to $w_i^*$. The following lemmas therefore relate the distances to a reference optimum $w_i^*$ before and after learning task $t$.

\begin{lemma}
\label{lemma:step_strong}
In the strongly convex case, the expected squared distance between the current model $w_t$ and the optimum $w_i^*$ of a previous task satisfies the following recursion, which involves the task gap $\|w_t^* - w_i^*\|^2$:
\begin{equation}
\label{eq:lemma_strong}
\begin{aligned}
&\mathbb{E}[\|w_t - w_i^*\|^2] \leq 4(1 - \gamma\mu)^n \mathbb{E}[\|w_{t-1} - w_i^*\|^2] \\
&\quad + \left(4(1 - \gamma\mu)^{n} + 2\right) \|w_t^* - w_i^*\|^2 + \frac{4\gamma\sigma_S^2}{\mu}.
\end{aligned}
\end{equation}
\end{lemma}

\begin{lemma}
\label{lemma:step_general}
In the general convex case, for any $\rho>1$, the expected squared distance between $w_t$ and a reference optimum $w_i^*$ satisfies
\begin{equation}
\label{eq:lemma_general}
\begin{aligned}
&\mathbb{E}[\|w_t - w_i^*\|^2] \leq \rho \mathbb{E}[\|w_{t-1} - w_i^*\|^2] \\
&\quad + \frac{4\rho}{\rho-1}\|w_t^* - w_i^*\|^2 + \frac{\rho}{\rho+1}\gamma^3 L n^2 \sigma_*^2 \\
&\quad - \frac{4\rho}{\rho+1}\gamma n \mathbb{E}[f_{t}(w_t) - f_{t}(w_t^*)].
\end{aligned}
\end{equation}
\end{lemma}

Both lemmas follow from Propositions~\ref{prop:intra_strong} and~\ref{prop:intra_general} together with Young's inequality. In the general convex case, we keep the two Young coefficients free and optimize them for a fixed product, which yields the tunable factor $\rho$. The proofs are given in the appendix.

Lemmas~\ref{lemma:step_strong} and~\ref{lemma:step_general} give one-step recursions for the distance to $w_i^*$. Because they are inequalities, unrolling them bounds $\mathbb{E}\|w_T-w_i^*\|^2$ and $\mathbb{E}\|w_i-w_i^*\|^2$ from above separately; however, the difference of two upper bounds is not an upper bound on the difference $\mathbb{E}\|w_T-w_i^*\|^2-\mathbb{E}\|w_i-w_i^*\|^2$ that defines $F_T$. We therefore unroll each recursion only from task $i$ to task $T$, which leaves the term $(\kappa^{T-i}-1)\,\mathbb{E}\|w_i-w_i^*\|^2$, where $\kappa$ denotes the one-step factor ($q$ or $\rho$ below). If $\kappa\ge1$, this coefficient is nonnegative and $\mathbb{E}\|w_i-w_i^*\|^2$ can be replaced by its upper bound; if $\kappa<1$, the coefficient is negative and the term can be dropped. This leads to our main results.

For the strongly convex case, we define the propagation factor $q = 4(1 - \gamma\mu)^n$, the variance constant $V = 4\gamma\sigma_S^2/\mu$, and, for task indices $1 \le i < j \le T$, the ordering coefficients
\begin{equation} \label{eq:ordering_coef}
c_{i,j} =
\begin{cases}
(q+2)(q^{T-i} - q^{j-i} + q^{T-j}), & q\ge 1,\\
(q+2)\,q^{T-j}, & q<1.
\end{cases}
\end{equation}
Note that $q<1$ if and only if $n>n_0:=\ln 4/\ln\frac{1}{1-\gamma\mu}$. The factor $4$ comes from the two Young inequalities in Lemma~\ref{lemma:step_strong}, so $q$ is the propagation factor of this bound, not the contraction factor $(1-\gamma\mu)^n<1$ of Shuffle SGD itself.

\begin{theorem} \label{thm1}
Under Assumptions~\ref{assump:smooth-convex}--\ref{assump:equal} in the strongly convex regime with stepsize $\gamma\le 1/L$, the expected forgetting after $T$ tasks satisfies the following.
(i) If $q\ge 1$,
\begin{equation} \label{eq:thm1_i}
\begin{split}
\mathbb{E}[F_T] &\leq \frac{1}{T-1} \sum_{i=1}^{T-1} \biggl\{ (q^T - q^i) \| w_i^*\|^2 \\
&\quad + \sum_{j>i}^{T} c_{i,j} \|w_i^* - w_j^*\|^2 + V\sum_{m=i}^{T-1} q^m \biggr\}.
\end{split}
\end{equation}
(ii) If $q<1$,
\begin{equation} \label{eq:thm1_ii}
\begin{split}
\mathbb{E}[F_T] &\leq \frac{1}{T-1} \sum_{i=1}^{T-1} \biggl\{ \sum_{j>i}^{T} c_{i,j} \|w_i^* - w_j^*\|^2 \\
&\quad + V\sum_{m=0}^{T-i-1} q^m \biggr\}.
\end{split}
\end{equation}
\end{theorem}

The two cases coincide at $q=1$, and $c_{i,j}\ge0$ in both: for $q\ge1$ because $q^{T-i}\ge q^{j-i}$, and for $q<1$ by definition. They differ only in how the subtracted term $\mathbb{E}\|w_i-w_i^*\|^2$ in $F_T$ is handled (Appendix~\ref{app:proof_thm1}).

For the general convex case, let $\eta,\theta>0$ be the two Young coefficients used to relate the optimum of task $t$ to the reference optimum, and set $\rho=(1+\eta)(1+\theta)>1$. Thus $\rho$ is the one-step growth factor of the general convex recursion; it is a free parameter of the analysis rather than an algorithmic hyperparameter, and the individual Young coefficients are optimized in the proof of Lemma~\ref{lemma:step_general}. For any $\rho>1$, we define the ordering coefficients
\begin{equation} \label{ocg}
g_{i,j} = \frac{4\rho}{\rho-1}\left(\rho^{T-i} - \rho^{j-i} + \rho^{T-j}\right).
\end{equation}

\begin{theorem} \label{thm2}
Under Assumptions~\ref{assump:smooth-convex}--\ref{assump:equal} in the general convex regime with stepsize $\gamma\le 1/(\sqrt{2}Ln)$, for any $\rho>1$ the expected forgetting after $T$ tasks satisfies
\begin{equation}
\begin{split}
\mathbb{E}[F_T]
&\leq \frac{1}{T-1}\sum_{i=1}^{T-1}\biggl\{
\sum_{j>i}^{T}g_{i,j}\|w_i^*-w_j^*\|^2 \\
&\quad +(\rho^T-\rho^i)\Bigl(\|w_i^*\|^2 + \frac{\rho\,\gamma^3Ln^2\sigma_*^2}{\rho^2-1}\Bigr)
\biggr\}.
\end{split}
\end{equation}
Since the bound holds for every $\rho>1$, with $g_{i,j}$ evaluated at the same $\rho$ by Eq.~\eqref{ocg}, one may take the infimum of the right-hand side over $\rho$.
\end{theorem}

Theorems~\ref{thm1} and~\ref{thm2} give explicit upper bounds on the expected forgetting in OCL; since the underlying single-task results are inequalities, they are bounds rather than exact expressions. We next examine three factors that govern these bounds.

(1) \emph{Data amount.} Much work on mitigating forgetting has addressed data imbalance across the task stream~\citep{kim2020imbalanced, wu2019large, zhao2020maintaining}, whereas it remains unclear how the amount of data within each task affects forgetting. Our bounds make the dependence on $n$ explicit.

(2) \emph{Task similarity.} The bounds depend on the gap $\|w_{i}^{*}-w_{j}^{*}\|^{2}$ between the optima of two tasks, which we use as a measure of task (dis)similarity: a larger gap means lower similarity. Understanding its effect helps explain interference between tasks and how similarity influences the stability of sequential learning.

(3) \emph{Task ordering.} For a fixed set of tasks, the order in which they are learned affects the bounds through the order-dependent coefficients $c_{i,j}$ and $g_{i,j}$. Understanding this effect is valuable because the bound may then be tightened simply by rearranging the task sequence, without changing the algorithm.

\section{Theoretical Interpretations}
\subsection{The impact of data amount}
\label{r4.1}
We first examine how the per-task data amount $n$ enters the bounds.

\textbf{1) Strongly convex objectives: exponential convergence of the task-gap part and linear growth of the variance part.} The bound of Theorem~\ref{thm1} depends on $n$ through two quantities: the propagation factor $q(n)=4(1-\gamma\mu)^n$, which decreases geometrically in $n$, and the variance constant $V(n)=4\gamma\sigma_S^2/\mu$, which grows with $n$ because the shuffling variance itself scales with $n$. Indeed, under Assumption~\ref{assump:smooth-convex}, $\frac{\gamma\mu n}{8}\sigma_*^2\le\sigma_S^2\le\frac{\gamma Ln}{4}\sigma_*^2$~\citep[Proposition~1]{mishchenko2021randomreshufflingsimpleanalysis}, so that
\begin{equation} \label{eq:V_n}
\frac{\gamma^2 n\sigma_*^2}{2}\;\le\;V(n)\;\le\;\frac{\gamma^2 L n\sigma_*^2}{\mu}.
\end{equation}
For $n\le n_0$, we have $q(n)\ge1$ and case~(i) applies; each of $q^T-q^i$, $c_{i,j}$, and $\sum_{m=i}^{T-1}q^m$ is nonnegative and nondecreasing in $q$, so apart from $V(n)$ the bound decreases as $n$ grows. For $n>n_0$, case~(ii) applies and the bound equals $G(n)+V(n)H(n)$, where
\begin{align*}
G(n)&=\frac{1}{T-1}\sum_{i=1}^{T-1}\sum_{j>i}^{T}(q+2)\,q^{T-j}\|w_i^*-w_j^*\|^2,\\
H(n)&=\frac{1}{T-1}\sum_{i=1}^{T-1}\sum_{m=0}^{T-i-1}q^m,
\end{align*}
with $q=q(n)$. Both $G$ and $H$ are increasing in $q$ and hence decreasing in $n$. As $n\to\infty$, $G(n)$ converges at the geometric rate $(1-\gamma\mu)^n$ to $\frac{2}{T-1}\sum_{i=1}^{T-1}\|w_i^*-w_T^*\|^2$ and $H(n)\to1$, whereas $V(n)$ grows linearly by Eq.~\eqref{eq:V_n}. Consequently, a larger $n$ tightens the task-gap part of the bound through $q$ but loosens its variance part, which grows linearly in $n$ because the shuffling variance accumulates over a longer epoch; for large $n$ the variance part dominates and the bound grows linearly in $n$. Whether the bound first decreases depends on the relative size of the two parts. If $\sigma_*=0$, i.e., all samples of a task share its optimum, then $V=0$ and the bound decreases monotonically in $n$.

\textbf{2) General convex objectives: an upper-bound implication.} In Theorem~\ref{thm2}, $n$ enters only through the accumulated variance term for any fixed $\rho>1$; the tunable Young coefficients change its constant but not its quadratic scaling in $n$. Note that Proposition~\ref{prop:intra_general} requires $\gamma\le 1/(\sqrt{2}Ln)$: for a fixed stepsize, the bound applies only to $n\le 1/(\sqrt{2}\gamma L)$, and in this range $\gamma^3Ln^2\sigma_*^2\le\gamma\sigma_*^2/(2L)$. Without the geometric contraction provided by strong convexity, the bound thus suggests that, within this range, a larger data amount can amplify the accumulated stochastic noise and loosen the bound. This is an implication of the upper bound rather than a tight characterization of every general convex instance.

\subsection{The impact of task similarity}
\label{r4.2}
In our analysis, forgetting is geometric: it measures how far the parameter trajectory is pulled away from the optimum of a past task while new tasks are learned. We quantify task similarity by the optimal-model gap $\Delta_{i,j} \triangleq \|w_i^* - w_j^*\|^2$; a smaller gap indicates more similar tasks with closer optima. This gap is an analytically convenient surrogate for task similarity rather than the only possible measure of relatedness; distributional distances such as the Wasserstein distance are left for future work. A key message of our bounds is that similarity does not enter the bounds uniformly: each gap is weighted by an ordering coefficient, which couples task similarity with task ordering.

In the strongly convex regime, the coefficients $c_{i,j}$ of Theorem~\ref{thm1} are nonnegative in both cases (Section~\ref{r4}). Hence a larger gap never decreases the bound; what depends on the positions of the two tasks is the weight of the gap. When $q<1$, this weight depends only on the position $j$ of the later task: a gap whose later task is learned last enters with weight $q+2$, whereas a gap whose later task is followed by $T-j$ further tasks is discounted by $q^{T-j}$, because each subsequent task scales the deviation induced by task $j$ by a factor $q$.

The general convex coefficients $g_{i,j}$ in Eq.~\eqref{ocg} are positive for all $i<j$ and any fixed $\rho>1$, because $\rho^{T-i}\ge\rho^{j-i}$ and $\rho^{T-j}>0$. Hence, here as well, a larger gap increases the bound and a smaller gap decreases it.

\subsection{The impact of task ordering}
\label{r4.3}
The ordering coefficients $c_{i,j}$ and $g_{i,j}$ depend only on the positions $i<j$ in the sequence, so their sum over all pairs of positions does not depend on which task is placed where. To isolate the effect of ordering, we assume that, for any tasks $i$ and $j$,
\begin{align*}
     &\|w_i^*-w_j^*\|^2 \\
     &\quad =
     \begin{cases}
        0.1, & \text{if task $i$ and $j$ are similar tasks;}\\
        1, & \text{if task $i$ and $j$ are dissimilar tasks.}
     \end{cases}
\end{align*}

Then the task-gap term of the bound equals $\sum_{i<j}c_{i,j}-0.9\sum_{(i,j)\in\mathcal{S}}c_{i,j}$, where $\mathcal{S}$ is the set of position pairs occupied by similar tasks and the first sum does not depend on the order; the same holds for $g_{i,j}$. Minimizing the task-gap term over task orders is therefore equivalent to maximizing the coefficients of the pairs of similar tasks, whose gap is smaller. We focus on this term; for $q>1$ and in the general convex case, the terms involving $\|w_i^*\|^2$ also depend on the order.

\textbf{Spacing similar tasks when $q<1$.} If $q<1$, then $c_{i,j}=(q+2)q^{T-j}$ increases with the position $j$ of the later task. Hence, for two similar tasks with the first one fixed, moving the second one later, i.e., increasing their distance, increases their coefficient and decreases the bound.

\textbf{Clustering similar tasks when $q>1$ or $\rho>1$.} If $q>1$ in the strongly convex case, or for any $\rho>1$ in the general convex case, the coefficients $c_{i,j}$ and $g_{i,j}$ in Eqs.~\eqref{eq:ordering_coef} and~\eqref{ocg} decrease as $j$ grows for fixed $i$, because the bases $q$ and $\rho$ exceed one. Increasing the distance between two similar tasks then decreases their coefficient and increases the task-gap term, so placing similar tasks close to each other is favored by this term.

Table~\ref{table.validate_1_groups_6} illustrates both cases for $T=6$. In each order, the two occurrences of $C_1$ are the two similar tasks and $C_2,\dots,C_5$ are dissimilar tasks; from top to bottom, the second similar task moves from position $2$ to position $6$, i.e., the distance between the two similar tasks increases from $1$ to $5$. The last two column give the coefficient of the similar pair up to a factor that does not depend on the order. When $q<1$, this coefficient grows from $q^4$ to $1$ as the two similar tasks move apart, so spacing them out minimizes the bound; when $\rho>1$, it shrinks from $\rho^5-\rho+\rho^4$ to $1$, so placing them next to each other minimizes the task-gap term.

\begin{table}[t]
\centering
\caption{Coefficient of the two similar tasks ($C_1$) for $T=6$, up to the factors $q+2$ and $\frac{4\rho}{\rho-1}$. For $q>1$, $c_{1,j}$ takes the form of $g_{1,j}$ with $\rho$ replaced by $q$.}
\label{table.validate_1_groups_6}
\setlength{\tabcolsep}{3.5pt}
\begin{tabular}{|c|c|c|c|}
\hline
\textbf{$d$} & \textbf{Order} & \textbf{$c_{1,j}$, $q<1$} & \textbf{$g_{1,j}$, $\rho>1$} \\ \hline
1 & $(C_1,C_1,C_2,C_3,C_4,C_5)$ & $q^4$ & $\rho^5-\rho+\rho^4$ \\ \hline
2 & $(C_1,C_2,C_1,C_3,C_4,C_5)$ & $q^3$ & $\rho^5-\rho^2+\rho^3$ \\ \hline
3 & $(C_1,C_2,C_3,C_1,C_4,C_5)$ & $q^2$ & $\rho^5-\rho^3+\rho^2$ \\ \hline
4 & $(C_1,C_2,C_3,C_4,C_1,C_5)$ & $q$ & $\rho^5-\rho^4+\rho$ \\ \hline
5 & $(C_1,C_2,C_3,C_4,C_5,C_1)$ & $1$ & $1$ \\ \hline
\end{tabular}
\end{table}

\section{Experimental Evaluation}

In this section, we examine whether the qualitative trends from the theory persist in practice. Next, we show MLP model experiment results. We report forgetting, the average accuracy (ACC), and the backward transfer (BWT) for the data amount, task similarity, and task ordering.

\textbf{Experimental setup.} We run online continual learning experiments with an MLP on Split-CIFAR-100~\citep{Krizhevsky2009LearningML} and PMNIST~\citep{lopezpaz2022gradientepisodicmemorycontinual}. The sequence contains $T=6$ tasks: each Split-CIFAR-100 task consists of ten CIFAR-100 classes, and each PMNIST task applies a fixed pixel permutation to MNIST images. For each task, the learner observes the task data once, trains without replay, and is then evaluated on all tasks seen so far. We use SGD with learning rate $0.01$, momentum $0.9$, batch size $128$, and no weight decay. All results are averaged over 5 random seeds, and further details are given in Appendix~\ref{sec:additional_exp_details}.

\begin{figure*}[ht]
    \centering
    \subfigure[\label{fig:nc-data}]{\includegraphics[width=0.32\textwidth]{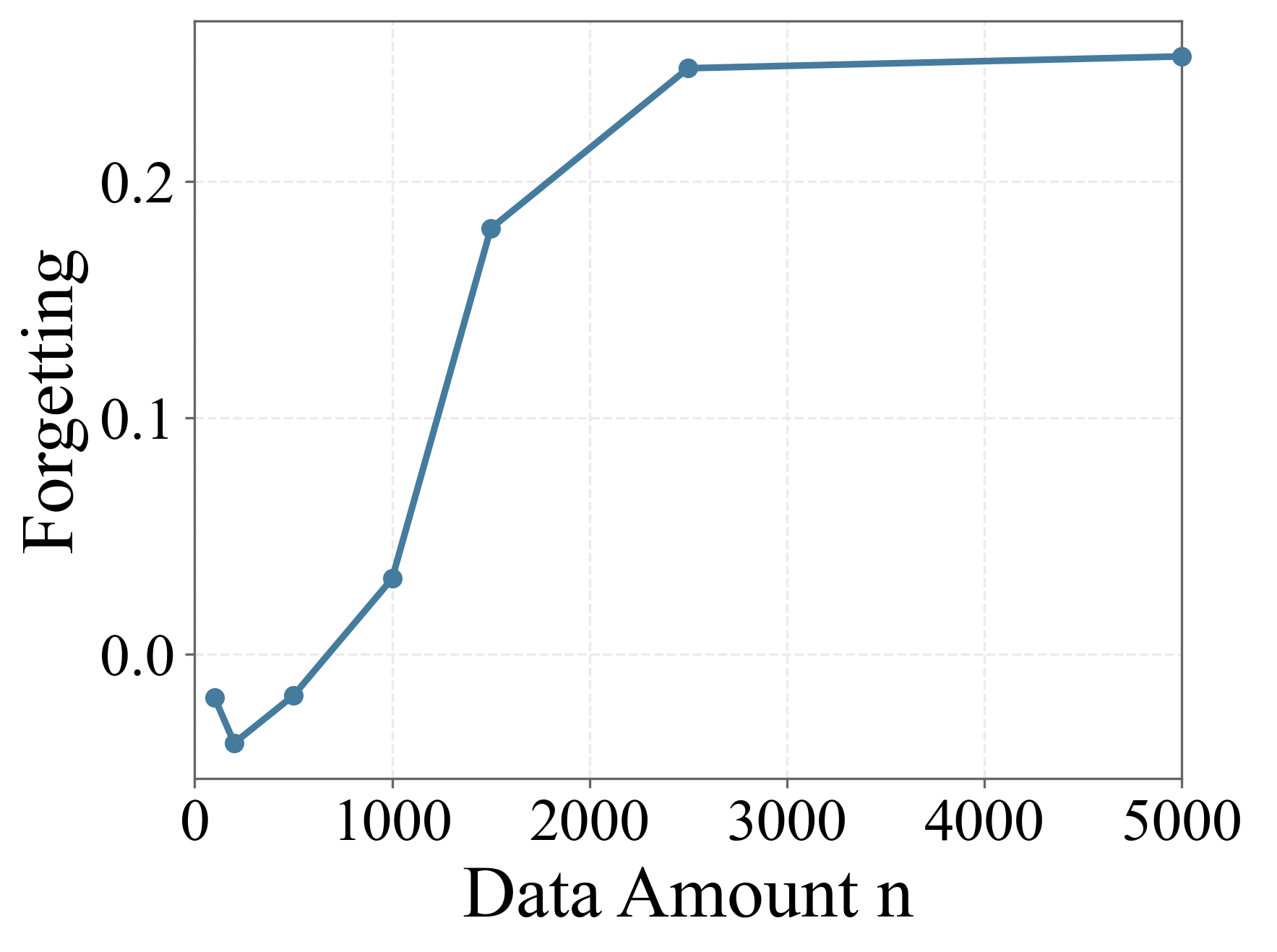}}
    \hfill
    \subfigure[\label{fig:nc-sim-task6}]{\includegraphics[width=0.32\textwidth]{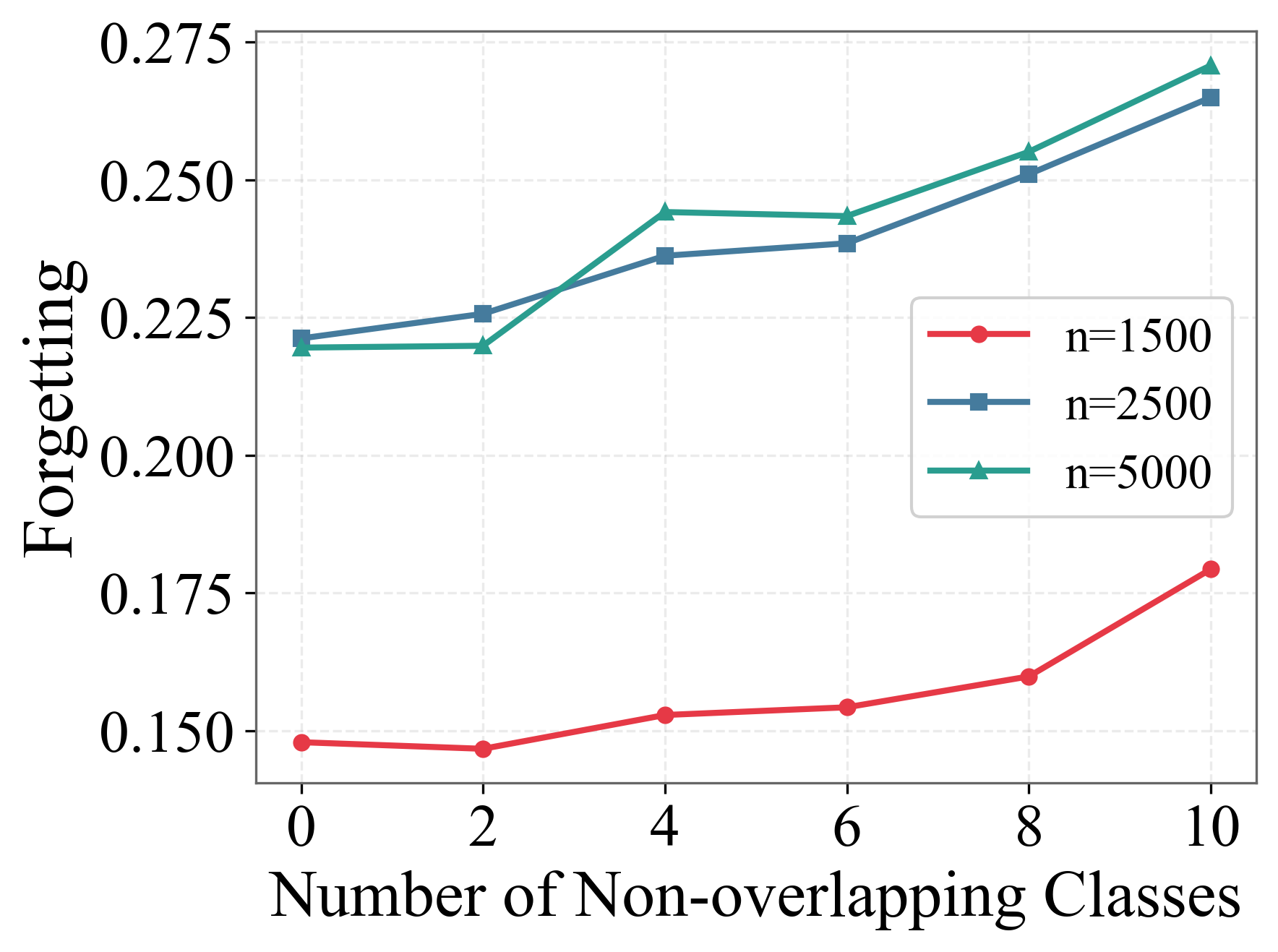}}
    \hfill
    \subfigure[\label{fig:nc-order}]{\includegraphics[width=0.32\textwidth]{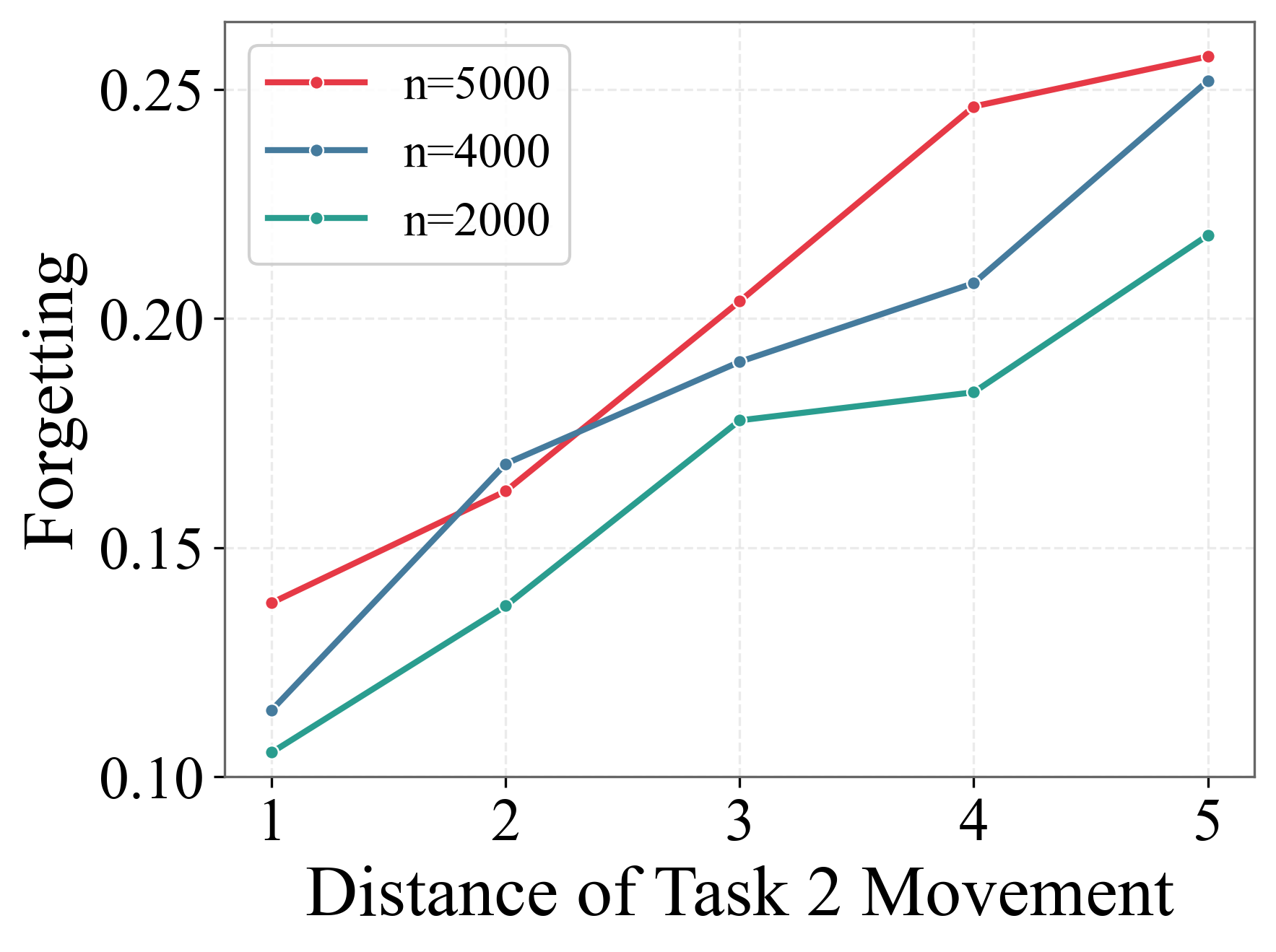}}
    \caption{Impact of data amount, task similarity, and task order. (a) MLP: forgetting increases with the amount of data. (b) MLP, with the target task learned last: forgetting increases with the number of non-overlapping classes, and a larger data amount leads to larger forgetting. (c) MLP: forgetting increases with the distance between the two similar tasks and with the amount of data. All results are averaged over 5 random seeds.}
    \label{fig:nonconvex_mlp}
\end{figure*}

\subsection{Data amount}

We first vary the per-task data amount $n$ while keeping the task sequence fixed. As shown in Figure~\ref{fig:nc-data}, the forgetting of the MLP increases sharply with $n$ and then plateaus, so more data per task does not automatically lead to better retention. Since the learner performs a single pass without replay, a larger $n$ also means more gradient updates on the current task, which can push the model further away from earlier tasks. This increase is consistent with the variance terms of our bounds: in the general convex case, the data amount enters only through a variance term that grows with $n$, and in the strongly convex case, the variance term grows linearly in $n$ and dominates the bound for large $n$ (Section~\ref{r4.1}).

\subsection{Task similarity}
We next study how task similarity affects forgetting. We fix Task~1 and vary a target task by changing the number of its classes that do not overlap with Task~1, so a larger value on the horizontal axis of Figure~\ref{fig:nc-sim-task6} corresponds to a less similar target task. The target task is learned last, after four intervening tasks.

Figure~\ref{fig:nc-sim-task6} shows that forgetting increases with the number of non-overlapping classes, and that a larger $n$ again leads to larger forgetting, in line with the data-amount effect above. This agrees with Section~\ref{r4.2}: the gap between two tasks enters the bounds with a nonnegative weight, so a less similar task never decreases the bound, and when $q<1$ this weight is largest when the later task is learned last.

\subsection{Task ordering}

\label{sec:task-ordering}

Finally, we evaluate the effect of task ordering. We construct a sequence with two similar tasks and four dissimilar tasks and move the second similar task farther away from Task~1. Figure~\ref{fig:nc-order} shows that forgetting increases with the distance between the two similar tasks, i.e., placing similar tasks close to each other reduces forgetting. This observation is consistent with our theoretical results: in the general convex case and in the strongly convex case with $q>1$, the task-gap terms of our bounds also favor placing similar tasks close to each other (Section~\ref{r4.3}).

\subsection{Accuracy and backward transfer}
\label{sec:fwt}

Besides forgetting, the performance of continual learning is commonly measured by the average accuracy (ACC) over all tasks and the backward transfer (BWT)~\citep{lopezpaz2022gradientepisodicmemorycontinual}, and recent work has related these metrics to each of the three factors in our analysis. Methods for imbalanced task streams adjust the effective contribution of data-rich and data-scarce tasks to facilitate knowledge transfer and mitigate forgetting~\citep{kim2020imbalanced, wu2019large, zhao2020maintaining, hong2024dynamically, wang2024rethinking, fang2024air}; empirical and theoretical studies show that the relationship between tasks strongly affects how much is forgotten~\citep{ramasesh2021anatomy, lee2021continual, doan2021theoretical}; and analyses of linear models show that the order of tasks changes the amount of forgetting~\citep{evron2022catastrophic, lin2023theory}. However, these works study each factor separately, and it remains unclear how the per-task data amount, task similarity, and task ordering affect ACC and BWT in the same online setting.

Existing methods for imbalanced streams do not isolate how the per-task data amount itself affects forgetting and forward knowledge transfer (FWT). Our bounds make explicit how the per-task data amount enters the sequence-dependent part of FWT.

Let $w_t^r$ denote the model trained on task $t$ alone from a random initialization. FWT is defined as
\begin{align}
\mathbb{E}\|w_t-w_t^*\|^2-\mathbb{E}\|w_t^r-w_t^*\|^2 .
\label{eq:fwt_def_main}
\end{align}
The second term does not depend on the preceding task sequence, so the sequence-dependent part of FWT is $\mathbb{E}\|w_t-w_t^*\|^2$. Unrolling Lemma~\ref{lemma:step_strong} from $w_0=0$ (Appendix~\ref{app:fwt}) gives
\begin{align}
\mathbb{E}\|w_t-w_t^*\|^2
&\le q^t\|w_t^*\|^2 \nonumber\\
&\quad +(q+2)\sum_{i=1}^{t-1}q^{t-i}\|w_i^*-w_t^*\|^2 \nonumber\\
&\quad +\frac{4\gamma\sigma_S^2}{\mu}\frac{1-q^t}{1-q},
\label{eq:fwt_strong_main}
\end{align}
where $q=4(1-\gamma\mu)^n$. This bounds only the first term of FWT; the scratch term $\mathbb{E}\|w_t^r-w_t^*\|^2$ also depends on $n$, so Eq.~\eqref{eq:fwt_strong_main} alone does not determine how $n$ affects FWT. Within the bound, the terms involving $\|w_t^*\|^2$ and the task gaps decrease with $n$ through $q$, whereas the variance term grows with $n$ through $\sigma_S^2$ (Eq.~\eqref{eq:V_n}). We therefore study the effect of the per-task data amount on transfer empirically. \citet{razdaibiedina2023progressivepromptscontinuallearning} also evaluate forward and backward transfer under different data amounts and observe the largest gains with less data.

We test this effect with MLPs on two standard continual-learning benchmarks, PMNIST~\citep{lopezpaz2022gradientepisodicmemorycontinual} and Split-CIFAR-100~\citep{Krizhevsky2009LearningML}. As shown in Table~\ref{tab:fwt_data_amount_mlp}, less data per task tends to yield higher accuracy and better backward transfer in these experiments.

\begin{table}[t]
\centering
\caption{Average final test accuracy (ACC) and backward transfer (BWT; the negative of forgetting, higher is better) over all tasks on PMNIST and Split-CIFAR-100, as the mean over 5 random seeds.}
\label{tab:fwt_data_amount_mlp}
\setlength{\tabcolsep}{4pt}
\begin{tabular}{cccccc}
\toprule
\multicolumn{1}{c}{\multirow{2}{*}{$n$}} & \multicolumn{2}{c}{PMNIST} &  & \multicolumn{2}{c}{Split CIFAR-100}\\ \cmidrule{2-3} \cmidrule{5-6}
\multicolumn{1}{c}{} & ACC(\%) & BWT(\%)   &  & ACC(\%) & BWT(\%)    \\ \midrule
1500                &
72.52   & 2.98 &  &
44.43   & -18.01 \\ \midrule
2500                  & 68.71   & -7.28 &  & 42.51   & -24.79   \\ \midrule
5000      & 57.89 & -15.56 & & 39.46 & -25.29 \\ \bottomrule
\end{tabular}
\end{table}

Table~\ref{tab:sim_acc_bwt} reports ACC and BWT for the task-similarity experiment with $n=2500$, where the dissimilarity $\delta$ is the fraction of the target task's classes that do not overlap with Task~1 on Split-CIFAR-100 and the fraction of its pixel positions reshuffled relative to Task~1 on PMNIST. On both datasets, ACC and BWT tend to decrease as $\delta$ increases. Thus a less similar task lowers the final accuracy and increases forgetting, in line with Section~\ref{r4.2}, where the gap between two tasks always enters the bounds with a nonnegative weight.

\begin{table}[!t]
\centering
\caption{ACC and BWT under different dissimilarities $\delta$ between the target task and Task~1 ($n=2500$).}
\label{tab:sim_acc_bwt}
\setlength{\tabcolsep}{4pt}
\begin{tabular}{cccccc}
\toprule
\multicolumn{1}{c}{\multirow{2}{*}{$\delta$}} & \multicolumn{2}{c}{PMNIST} &  & \multicolumn{2}{c}{Split CIFAR-100}\\ \cmidrule{2-3} \cmidrule{5-6}
\multicolumn{1}{c}{} & ACC(\%) & BWT(\%) &  & ACC(\%) & BWT(\%) \\ \midrule
0\%   & 75.13 & 0.64  &  & 17.71 & -22.35 \\ \midrule
40\%  & 71.03 & -4.51 &  & 16.90 & -23.68 \\ \midrule
100\% & 70.20 & -5.46 &  & 14.32 & -26.53 \\ \bottomrule
\end{tabular}
\end{table}

Table~\ref{tab:order_acc_bwt} reports ACC and BWT for the task-ordering experiment with $n=2000$, in a variant that follows our analysis more closely: the two similar tasks share the same classes on Split-CIFAR-100 and the same permutation on PMNIST, but use disjoint training images. On both datasets, ACC and BWT tend to decrease as the second similar task is placed farther from the first. Placing similar tasks close to each other thus improves both metrics, consistent with the ordering favored by the task-gap terms of our bounds in the general convex case and in the strongly convex case with $q>1$ (Section~\ref{r4.3}).

\begin{table}[!tb]
\centering
\caption{ACC and BWT under different distances $d$ between the two similar tasks ($n=2000$).}
\label{tab:order_acc_bwt}
\setlength{\tabcolsep}{4pt}
\begin{tabular}{cccccc}
\toprule
\multicolumn{1}{c}{\multirow{2}{*}{$d$}} & \multicolumn{2}{c}{PMNIST} &  & \multicolumn{2}{c}{Split CIFAR-100}\\ \cmidrule{2-3} \cmidrule{5-6}
\multicolumn{1}{c}{} & ACC(\%) & BWT(\%) &  & ACC(\%) & BWT(\%) \\ \midrule
1 & 75.42 & 5.86 &  & 23.59 & 2.02  \\ \midrule
3 & 74.24 & 5.78 &  & 23.24 & 1.81  \\ \midrule
5 & 73.62 & 5.28 &  & 21.74 & -0.49 \\ \bottomrule
\end{tabular}
\end{table}

\section{Conclusion}

We developed an explicit theoretical framework for catastrophic forgetting in online continual learning under strongly convex and general convex objectives. We derived upper bounds on the expected forgetting of single-pass Shuffle SGD and showed how data amount, task similarity, and task ordering enter them, with effects that depend on the objective class. The empirical results are consistent with the qualitative role of these factors, suggesting that the theory provides useful guidance for understanding OCL behavior. Future work includes extending the analysis to structured non-convex settings, deriving matching lower bounds, and turning the ordering coefficients into practical task-scheduling algorithms.

\bibliography{nips2026-template}

\clearpage
\appendix
\thispagestyle{empty}

\onecolumn
\aistatstitle{Formalizing Forgetting: Theory of Forgetting for Online Continual Learning: \\
Supplementary Materials}

\section{Key Notation}
\label{app:notation}

\bgroup
\def\arraystretch{1.5}
\begin{longtable}{@{}p{1.25in}p{4.5in}}
$\displaystyle T,\ \mathcal{T}$ & Number of tasks and the task index set $\mathcal{T}=\{1,\dots,T\}$. \\
$\displaystyle S_t,\ \mathcal{D}_t$ & Training set of task $t$, with $n$ samples drawn i.i.d.\ from the distribution $\mathcal{D}_t$. \\
$\displaystyle f_t,\ f_{t,i}$ & Finite-sum objective of task $t$ and its $i$-th component, $f_t=\frac{1}{n}\sum_{i=1}^n f_{t,i}$. \\
$\displaystyle w_t^*$ & A fixed minimizer of $f_t$. \\
$\displaystyle w_t,\ w_0$ & Parameter obtained after training on task $t$; initialization $w_0=0$. \\
$\displaystyle \pi_t$ & Random permutation used by Shuffle SGD on task $t$. \\
$\displaystyle K_i(w),\ F_t$ & Model error $\|w-w_i^*\|^2$ on task $i$, and forgetting after learning task $t$. \\
$\displaystyle \gamma$ & Stepsize of Shuffle SGD. \\
$\displaystyle L$ & Smoothness constant of every component $f_{t,i}$. \\
$\displaystyle \mu$ & Strong convexity constant of every component $f_{t,i}$ (strongly convex case). \\
$\displaystyle n$ & Number of samples per task, i.e., the number of Shuffle-SGD steps in one epoch. \\
$\displaystyle \sigma_*$ & Variance at the optimum: $\displaystyle \sigma_*^2 := \frac{1}{n} \sum_{i=1}^n \| \nabla f_{t,i}(w_t^*) \|^2$. \\
$\displaystyle \sigma_{\text{S}}$ & Shuffling variance~\citep{mishchenko2021randomreshufflingsimpleanalysis}: $\displaystyle \sigma_{\text{S}}^2 := \max_{k=1,\dots,n-1} \frac{1}{\gamma} \mathbb{E} \left[ D_{f_{t,\pi_t(k)}}\!\left(w_{t,*}^{k}, w_t^*\right) \right]$, where $w_{t,*}^{k} := w_t^* - \gamma\sum_{j=0}^{k-1}\nabla f_{t,\pi_t(j)}(w_t^*)$ and $D_f(x,y):=f(x)-f(y)-\langle\nabla f(y),x-y\rangle$ is the Bregman divergence. \\
$\displaystyle q,\ V$ & Propagation factor $q=4(1-\gamma\mu)^n$ of the bound in Theorem~\ref{thm1} and variance constant $V=4\gamma\sigma_S^2/\mu$ (strongly convex case). \\
$\displaystyle n_0$ & Threshold $n_0=\ln 4/\ln\frac{1}{1-\gamma\mu}$, so that $q<1$ if and only if $n>n_0$. \\
$\displaystyle \eta,\ \theta,\ \rho$ & Young coefficients of the general convex analysis and the one-step growth factor $\rho=(1+\eta)(1+\theta)$ of the general convex recursion, a free parameter $\rho>1$ of the analysis. \\
$\displaystyle c_{i,j},\ g_{i,j}$ & Ordering coefficients of the task gap between positions $i<j$ in Theorems~\ref{thm1} and~\ref{thm2}. \\
$\displaystyle \Delta_{i,j}$ & Task gap $\|w_i^*-w_j^*\|^2$, used as a measure of task dissimilarity. \\
$\displaystyle \mathcal{S}$ & Set of position pairs occupied by similar tasks (Section~\ref{r4.3}). \\
$\displaystyle A_{t,i}$ & Test accuracy on task $i$ after training on task $t$. \\
ACC, BWT & Average accuracy and backward transfer (Appendix~\ref{sec:additional_exp_details}). \\
$\displaystyle w_t^r$ & Model trained on task $t$ alone from a random initialization, used to define FWT. \\
$\displaystyle \delta,\ d$ & Dissimilarity between the target task and Task~1, and distance between the two similar tasks, in Tables~\ref{tab:sim_acc_bwt} and~\ref{tab:order_acc_bwt}. \\
\end{longtable}
\egroup
\vspace{0.25cm}

\section{Proof of Lemma \texorpdfstring{\ref{lemma:step_strong}}{1}}

For strongly convex objectives and $\gamma\le 1/L$, Theorem~1 of \citet{mishchenko2021randomreshufflingsimpleanalysis} gives
\begin{equation}
\label{proof1}
\mathbb{E}[\|w_t - w_t^*\|^2] \le (1-\gamma\mu)^{n}\, \mathbb{E}[\|w_{t-1} - w_t^*\|^2] + \frac{2\gamma\sigma_{\text{S}}^2}{\mu}
\end{equation}

By Young's inequality,
\begin{align*}
\|w_t - w_i^*\|^2 &= \|(w_t - w_t^*) + (w_t^* - w_i^*)\|^2 \\
&\le 2 \| w_t - w_t^* \| ^2 + 2 \| w_t^* - w_i^* \| ^2
\end{align*}
Taking expectations gives
\begin{equation}
\label{proof2}
\mathbb{E}[\|w_t - w_i^*\|^2] \le 2\mathbb{E}[\|w_t - w_t^*\|^2] + 2\|w_t^* - w_i^*\|^2
\end{equation}

Substituting Eq.~\eqref{proof1} into Eq.~\eqref{proof2} gives
\begin{equation}
\label{proof3}
\mathbb{E}[\|w_t - w_i^*\|^2] \le 2\left( (1-\gamma\mu)^{n}\, \mathbb{E}[\|w_{t-1} - w_t^*\|^2] + \frac{2\gamma\sigma_{\text{S}}^2}{\mu} \right) + 2\|w_t^* - w_i^*\|^2
\end{equation}

To bound $\|w_{t-1} - w_t^*\|^2$, we use Young's inequality again:
\begin{align*}
\|w_{t-1} - w_t^*\|^2 &= \|(w_{t-1} - w_i^*) + (w_i^* - w_t^*)\|^2 \\
&\le 2\|w_{t-1} - w_i^*\|^2 + 2\|w_i^* - w_t^*\|^2
\end{align*}

Substituting this bound into Eq.~\eqref{proof3} proves the lemma:
\begin{align}
\label{proof4}
\mathbb{E}[\|w_t - w_i^*\|^2] &\le 4(1-\gamma\mu)^{n}\, \mathbb{E}[\|w_{t-1} - w_i^*\|^2] + \left( 4(1-\gamma\mu)^{n} + 2 \right) \|w_t^* - w_i^*\|^2 + \frac{4\gamma\sigma_{\text{S}}^2}{\mu}
\end{align}

\section{Proof of Theorem~\ref{thm1}}
\label{app:proof_thm1}

Let $q = 4(1-\gamma\mu)^n$ and $V = 4\gamma\sigma_S^2/\mu$. By Lemma~\ref{lemma:step_strong}, for every reference task $i$ and every $t\ge1$,
\begin{equation}
\label{proof5}
\mathbb{E}\|w_t-w_i^*\|^2 \le q\,\mathbb{E}\|w_{t-1}-w_i^*\|^2 + (q+2)\|w_t^*-w_i^*\|^2 + V.
\end{equation}
Since Eq.~\eqref{proof5} is an inequality, unrolling it yields upper bounds only. In particular, bounding $\mathbb{E}\|w_T-w_i^*\|^2$ and $\mathbb{E}\|w_i-w_i^*\|^2$ from above separately and subtracting the two bounds does not give an upper bound on $\mathbb{E}\|w_T-w_i^*\|^2-\mathbb{E}\|w_i-w_i^*\|^2$, because the subtracted quantity would require a lower bound. We therefore proceed as follows.

\paragraph{Step 1: unrolling from task $i$ to task $T$.} All coefficients in Eq.~\eqref{proof5} are nonnegative, so applying it for $t=T,T-1,\ldots,i+1$ gives
\begin{equation}
\label{proof6}
\mathbb{E}\|w_T-w_i^*\|^2 \le q^{T-i}\,\mathbb{E}\|w_i-w_i^*\|^2 + (q+2)\sum_{k=i+1}^{T}q^{T-k}\|w_k^*-w_i^*\|^2 + V\sum_{m=0}^{T-i-1}q^m .
\end{equation}
Subtracting $\mathbb{E}\|w_i-w_i^*\|^2$ from both sides, which is exact, yields
\begin{equation}
\label{proof6b}
\mathbb{E}\|w_T-w_i^*\|^2-\mathbb{E}\|w_i-w_i^*\|^2 \le (q^{T-i}-1)\,\mathbb{E}\|w_i-w_i^*\|^2 + R_i,
\end{equation}
where $R_i := (q+2)\sum_{k=i+1}^{T}q^{T-k}\|w_k^*-w_i^*\|^2 + V\sum_{m=0}^{T-i-1}q^m$.

\paragraph{Step 2, case $q\ge1$.} Then $q^{T-i}-1\ge0$, so the first term on the right-hand side of Eq.~\eqref{proof6b} can be bounded from above with an upper bound on $\mathbb{E}\|w_i-w_i^*\|^2$. Unrolling Eq.~\eqref{proof5} from $t=1$ to $t=i$ with $w_0=0$ gives
\begin{equation}
\label{proof6c}
\mathbb{E}\|w_i-w_i^*\|^2 \le q^{i}\|w_i^*\|^2 + (q+2)\sum_{k=1}^{i}q^{i-k}\|w_k^*-w_i^*\|^2 + V\sum_{m=0}^{i-1}q^m .
\end{equation}
Plugging Eq.~\eqref{proof6c} into Eq.~\eqref{proof6b} and using $(q^{T-i}-1)\sum_{m=0}^{i-1}q^m+\sum_{m=0}^{T-i-1}q^m=\sum_{m=i}^{T-1}q^m$, we obtain
\begin{align*}
\mathbb{E}\|w_T-w_i^*\|^2-\mathbb{E}\|w_i-w_i^*\|^2
&\le (q^T-q^i)\|w_i^*\|^2 + (q+2)\sum_{k=1}^{i}\left(q^{T-k}-q^{i-k}\right)\|w_k^*-w_i^*\|^2 \\
&\quad + (q+2)\sum_{k=i+1}^{T}q^{T-k}\|w_k^*-w_i^*\|^2 + V\sum_{m=i}^{T-1}q^m .
\end{align*}
The term $k=i$ vanishes. After averaging over $i=1,\ldots,T-1$, a pair $i<j$ receives the coefficient $(q+2)q^{T-j}$ from the forgetting of task $i$ (term $k=j$ of the second sum) and the coefficient $(q+2)(q^{T-i}-q^{j-i})$ from the forgetting of task $j$ (term $k=i$ of the first sum, present for $j\le T-1$; for $j=T$ this coefficient equals zero). Their sum is $c_{i,j}=(q+2)(q^{T-i}-q^{j-i}+q^{T-j})$, which proves case~(i).

\paragraph{Step 2, case $q<1$.} Then $q^{T-i}-1<0$ while $\mathbb{E}\|w_i-w_i^*\|^2\ge0$, so the first term on the right-hand side of Eq.~\eqref{proof6b} is nonpositive and can be dropped:
\begin{equation*}
\mathbb{E}\|w_T-w_i^*\|^2-\mathbb{E}\|w_i-w_i^*\|^2 \le (q+2)\sum_{j=i+1}^{T}q^{T-j}\|w_j^*-w_i^*\|^2 + V\sum_{m=0}^{T-i-1}q^m .
\end{equation*}
Note that Eq.~\eqref{proof6c} cannot be used in this case, since multiplying an upper bound by the negative factor $q^{T-i}-1$ does not give an upper bound. Averaging over $i=1,\ldots,T-1$ proves case~(ii) with $c_{i,j}=(q+2)q^{T-j}$. \hfill$\square$

\section{Proof of Lemma \texorpdfstring{\ref{lemma:step_general}}{2}}

For general convex objectives and $\gamma\le 1/(\sqrt{2}Ln)$, the per-epoch analysis of Shuffle SGD in \citet{mishchenko2021randomreshufflingsimpleanalysis} gives
\begin{equation}
\label{proof7}
\mathbb{E}[\|w_t-w_t^{*}\|^2] \le \mathbb{E}[\|w_{t-1}-w_t^{*}\|^2] - 2\gamma n\mathbb{E}[f_{t}(w_t)-f_{t}(w_t^*)] + \frac{\gamma^{3}Ln^{2}\sigma_{*}^{2}}{2}
\end{equation}

For any $\eta,\theta>0$, Young's inequality gives
\begin{align}
\mathbb{E}\|w_t-w_i^*\|^2
&\le (1+\eta)\mathbb{E}\|w_t-w_t^*\|^2
+\left(1+\frac{1}{\eta}\right)\|w_t^*-w_i^*\|^2,\\
\mathbb{E}\|w_{t-1}-w_t^*\|^2
&\le (1+\theta)\mathbb{E}\|w_{t-1}-w_i^*\|^2
+\left(1+\frac{1}{\theta}\right)\|w_t^*-w_i^*\|^2.
\end{align}
Substituting the second inequality and~\eqref{proof7} into the first one yields
\begin{align}
\mathbb{E}\|w_t-w_i^*\|^2
&\le (1+\eta)(1+\theta)\mathbb{E}\|w_{t-1}-w_i^*\|^2 \nonumber\\
&\quad +\left[(1+\eta)\left(1+\frac{1}{\theta}\right)+\left(1+\frac{1}{\eta}\right)\right]\|w_t^*-w_i^*\|^2 \nonumber\\
&\quad -2\gamma n(1+\eta)\mathbb{E}[f_{t}(w_t)-f_{t}(w_t^*)]
+(1+\eta)\frac{\gamma^3Ln^2\sigma_*^2}{2}.
\label{proof10_pre}
\end{align}
Set $(1+\eta)(1+\theta)=\rho>1$. As a function of $\eta\in(0,\rho-1)$, the coefficient of the task gap diverges as $\eta\to0^+$ and as $\eta\to(\rho-1)^-$; hence its unique stationary point is the global minimizer,
\[
\eta=\frac{\rho-1}{\rho+1},\qquad \theta=\frac{\rho-1}{2},
\]
and the minimized coefficient is $\frac{4\rho}{\rho-1}$. Substituting these values into~\eqref{proof10_pre} proves Lemma~\ref{lemma:step_general}.

\section{Proof of Theorem~\ref{thm2}}

Fix $\rho>1$ and drop the nonpositive descent term in Lemma~\ref{lemma:step_general}. Let
\[
B_\rho=\frac{4\rho}{\rho-1},\qquad
V_\rho=\frac{\rho}{\rho+1}\gamma^3Ln^2\sigma_*^2 .
\]
Then
\[
\mathbb{E}\|w_t-w_i^*\|^2
\le \rho\,\mathbb{E}\|w_{t-1}-w_i^*\|^2+B_\rho\|w_t^*-w_i^*\|^2+V_\rho.
\]
Propagating this recursion from $i$ to $T$ gives
\begin{align}
\mathbb{E}\|w_T-w_i^*\|^2
&\le \rho^{T-i}\mathbb{E}\|w_i-w_i^*\|^2
+B_\rho\sum_{k=i+1}^{T}\rho^{T-k}\|w_k^*-w_i^*\|^2
+V_\rho\sum_{k=i+1}^{T}\rho^{T-k}.
\label{eq:direct_gc}
\end{align}
Equation~\eqref{eq:direct_gc} implies
\begin{align}
\mathbb{E}\|w_T-w_i^*\|^2-\mathbb{E}\|w_i-w_i^*\|^2
&\le(\rho^{T-i}-1)\mathbb{E}\|w_i-w_i^*\|^2 \nonumber\\
&\quad +B_\rho\sum_{k=i+1}^{T}\rho^{T-k}\|w_k^*-w_i^*\|^2
+V_\rho\sum_{k=i+1}^{T}\rho^{T-k}.
\label{eq:gc_before_di}
\end{align}
Since $\rho^{T-i}-1\ge0$, the first term on the right-hand side can be bounded from above by unrolling the same recursion from $0$ to $i$ with $w_0=0$:
\[
\mathbb{E}\|w_i-w_i^*\|^2
\le \rho^i\|w_i^*\|^2
+B_\rho\sum_{k=1}^{i}\rho^{i-k}\|w_k^*-w_i^*\|^2
+V_\rho\sum_{k=1}^{i}\rho^{i-k}.
\]
Substituting this into~\eqref{eq:gc_before_di} yields
\begin{align}
\mathbb{E}\|w_T-w_i^*\|^2-\mathbb{E}\|w_i-w_i^*\|^2
&\le(\rho^T-\rho^i)\|w_i^*\|^2 \nonumber\\
&\quad +B_\rho\left[\sum_{k=1}^{i}(\rho^{T-k}-\rho^{i-k})\|w_k^*-w_i^*\|^2
+\sum_{k=i+1}^{T}\rho^{T-k}\|w_k^*-w_i^*\|^2\right]\nonumber\\
&\quad +\frac{V_\rho}{\rho-1}(\rho^T-\rho^i).
\end{align}
Collecting the task-gap terms pairwise gives
\[
g_{i,j}
=B_\rho(\rho^{T-i}-\rho^{j-i}+\rho^{T-j}).
\]
Averaging over $i=1,\ldots,T-1$ gives Theorem~\ref{thm2}; since the argument holds for every $\rho>1$, we take the infimum over $\rho$.

\section{Additional Experimental Details}
\label{sec:additional_exp_details}

\paragraph{Models.}
For Split-CIFAR-100, the MLP takes flattened $32\times32\times3$ images as input and uses two hidden layers, $3072\to512\to256\to100$, with ReLU activations. For PMNIST, the MLP takes flattened $28\times28$ images and uses $784\to512\to256\to10$, also with ReLU activations.

\paragraph{Task construction.}
For Split-CIFAR-100, each task is formed by selecting ten CIFAR-100 classes, and the learner uses a shared 100-way output head. In the data-amount experiment, the task sequence is fixed while the number $n$ of samples per task is varied. In the task-similarity experiment, Task~1 is fixed and the target task is constructed by changing the number of its classes that do not overlap with Task~1. In the task-ordering experiment, the sequence contains two similar tasks and four dissimilar tasks, and we vary the position of the second similar task. For PMNIST, each task is a permuted MNIST task with a fixed pixel permutation and a shared 10-way output head. In the task-similarity experiment, the target task uses the permutation of Task~1 with a fraction $\delta$ of its positions reshuffled; in the task-ordering experiment, the two similar tasks use the same permutation and draw their training images from two disjoint halves of the MNIST training set. In the task-similarity and task-ordering experiments, the accuracy on a Split-CIFAR-100 task is computed over its ten classes.

\paragraph{Training protocol and resources.}
All experiments follow the online single-pass setting: after a task is observed once, its samples are not replayed. Models are trained with mini-batch SGD using step size (learning rate) $\gamma=0.01$, momentum $0.9$, batch size $128$, and no weight decay; the task-ordering experiment in Table~\ref{tab:order_acc_bwt} uses the same setting without momentum. Experiments were run on a single workstation with an NVIDIA GeForce RTX 3060 Laptop GPU. The runtime depends on data caching and device availability.

\paragraph{Evaluation metrics.}
We evaluate average accuracy (ACC) and backward transfer (BWT)~\citep{lopezpaz2022gradientepisodicmemorycontinual}, the empirical counterparts of the quantities in our analysis. Formally, they are defined as
\begin{align*}
\mathrm{ACC}=\frac{1}{T}\sum_{i=1}^{T}A_{T,i}, \qquad
\mathrm{BWT}=\frac{1}{T-1}\sum_{i=1}^{T-1}(A_{T,i}-A_{i,i}),
\end{align*}
where $A_{t,i}$ denotes the accuracy of the model on task $i$ after completing training on task $t$, and $T$ is the total number of tasks.

\paragraph{Hyperparameter choice.}
We use the same optimizer and batch size in all experiments to isolate the effects of data amount, task similarity, and task ordering. The step size $\gamma=0.01$ with momentum $0.9$ is a standard, stable SGD setting for these vision benchmarks, batch size $128$ gives stable mini-batch estimates while fitting easily on the GPU, and weight decay is set to zero to avoid introducing an additional regularization factor into the forgetting trends.

\section{Forward Transfer Derivations}
\label{app:fwt}

We derive Eq.~\eqref{eq:fwt_strong_main} used in Section~\ref{sec:fwt}. For task $t$, let $w_t^r$ be the model trained on task $t$ alone from a random initialization. The FWT error gap is
\[
{\rm FWT}_t
=\mathbb{E}\|w_t-w_t^*\|^2-\mathbb{E}\|w_t^r-w_t^*\|^2 .
\]
The scratch term $\mathbb{E}\|w_t^r-w_t^*\|^2$ does not depend on the preceding task sequence, so the only sequence-dependent term is $\mathbb{E}\|w_t-w_t^*\|^2$.

Let
\[
q=4(1-\gamma\mu)^n,\qquad V=\frac{4\gamma\sigma_S^2}{\mu}.
\]
We first derive a one-task recursion with reference optimum $w_t^*$. For the update on task $k$, Proposition~\ref{prop:intra_strong} gives
\begin{equation}
\mathbb{E}\|w_k-w_k^*\|^2
\le (1-\gamma\mu)^n\mathbb{E}\|w_{k-1}-w_k^*\|^2
+\frac{2\gamma\sigma_S^2}{\mu}.
\label{eq:fwt_appendix_shuffle}
\end{equation}
By Young's inequality,
\begin{align}
\mathbb{E}\|w_k-w_t^*\|^2
&\le 2\mathbb{E}\|w_k-w_k^*\|^2
+2\|w_k^*-w_t^*\|^2, \label{eq:fwt_appendix_young_1}\\
\mathbb{E}\|w_{k-1}-w_k^*\|^2
&\le 2\mathbb{E}\|w_{k-1}-w_t^*\|^2
+2\|w_k^*-w_t^*\|^2. \label{eq:fwt_appendix_young_2}
\end{align}
Substituting Eq.~\eqref{eq:fwt_appendix_shuffle} and Eq.~\eqref{eq:fwt_appendix_young_2} into Eq.~\eqref{eq:fwt_appendix_young_1} yields, for $1\le k\le t$,
\begin{align}
\mathbb{E}\|w_k-w_t^*\|^2
&\le q\,\mathbb{E}\|w_{k-1}-w_t^*\|^2
+(q+2)\|w_k^*-w_t^*\|^2+V .
\label{eq:fwt_appendix_step}
\end{align}
We now unroll this recursion from $k=1$ to $k=t$. For $k=1$,
\[
\mathbb{E}\|w_1-w_t^*\|^2
\le q\|w_0-w_t^*\|^2
+(q+2)\|w_1^*-w_t^*\|^2+V .
\]
For $k=2$,
\[
\mathbb{E}\|w_2-w_t^*\|^2
\le q^2\|w_0-w_t^*\|^2
+(q+2)\sum_{i=1}^{2}q^{2-i}\|w_i^*-w_t^*\|^2
+V\sum_{i=1}^{2}q^{2-i}.
\]
Repeating the same substitution yields the general form
\begin{align}
\mathbb{E}\|w_t-w_t^*\|^2
&\le q^t\|w_0-w_t^*\|^2
+(q+2)\sum_{i=1}^{t}q^{t-i}\|w_i^*-w_t^*\|^2
+V\sum_{i=1}^{t}q^{t-i}.
\label{eq:fwt_appendix_unroll}
\end{align}
Since $w_0=0$, we have $\|w_0-w_t^*\|^2=\|w_t^*\|^2$. Moreover, the term with $i=t$ in the task-gap sum is zero because $\|w_t^*-w_t^*\|^2=0$. Therefore,
\begin{align}
\mathbb{E}\|w_t-w_t^*\|^2
&\le q^t\|w_t^*\|^2
+(q+2)\sum_{i=1}^{t-1}q^{t-i}\|w_i^*-w_t^*\|^2
+V\sum_{i=1}^{t}q^{t-i} \nonumber\\
&= q^t\|w_t^*\|^2
+(q+2)\sum_{i=1}^{t-1}q^{t-i}\|w_i^*-w_t^*\|^2
+\frac{4\gamma\sigma_S^2}{\mu}\frac{1-q^t}{1-q},
\label{eq:fwt_appendix_final}
\end{align}

\section{Limitations}
\label{limi}
Our framework is a step toward formalizing forgetting in OCL, but several limitations remain. First, although the bounds capture the qualitative behavior of forgetting under convex objectives, extending them to non-convex settings is challenging; structured non-convex models that satisfy the Polyak--{\L}ojasiewicz condition, or input-convex architectures, may help bridge this gap. Second, the optimal-model distance characterizes task similarity from the perspective of the optimizer geometry, whereas capturing data-level relatedness, e.g., through the Wasserstein distance, remains open. Third, our results are upper bounds without matching lower bounds or tightness guarantees; such lower bounds would clarify to what extent the identified factors are necessary, rather than merely sufficient, for forgetting. In particular, when $q<1$ in Theorem~\ref{thm1}, the nonpositive term $(q^{T-i}-1)\,\mathbb{E}\|w_i-w_i^*\|^2$ is dropped, and retaining it would require a lower bound on $\mathbb{E}\|w_i-w_i^*\|^2$. Moreover, the propagation factor $q$ depends on the Young coefficients used in Lemma~\ref{lemma:step_strong}, so the ordering preference for $q>1$ reflects this particular bound rather than a property of Shuffle SGD. Finally, although the ordering coefficients $c_{i,j}$ and $g_{i,j}$ reveal how the task arrangement influences forgetting, turning them into practical scheduling algorithms that automatically determine good task orders is a promising direction for future work.

\end{document}